# KNOWLEDGE CAPTURE, ADAPTAION AND COMPOSITION (KCAC): A FRAMEWORK FOR CROSS-TASK CURRICULUM LEARNING IN ROBOTIC MANIPULATION

**Xinrui Wang**
Dept. of Aerospace & Mechanical Engineering
University of Southern California
Los Angeles, USA
xinruiw@usc.edu

**Yan Jin***
Dept. of Aerospace & Mechanical Engineering
University of Southern California
Los Angeles, USA
yjin@usc.edu
(*corresponding author)

**ABSTRACT**

*Reinforcement learning (RL) has demonstrated remarkable potential in robotic manipulation but faces challenges in sample inefficiency and lack of interpretability, limiting its applicability in real-world scenarios. Enabling the agent to gain a deeper understanding and adapt more efficiently to diverse working scenarios is crucial, and strategic knowledge utilization is a key factor in this process. This paper proposes a Knowledge Capture, Adaptation, and Composition (KCAC) framework to systematically integrate knowledge transfer into RL through cross-task curriculum learning. KCAC is evaluated using a two-block stacking task in the CausalWorld benchmark, a complex robotic manipulation environment. To our knowledge, existing RL approaches fail to solve this task effectively, reflecting deficiencies in knowledge capture. In this work, we redesign the benchmark reward function by removing rigid constraints and strict ordering, allowing the agent to maximize total rewards concurrently and enabling flexible task completion. Further, we define two self-designed sub-tasks and implement a structured cross-task curriculum to facilitate efficient learning. As a result, our KCAC approach achieves a 40% reduction in training time while improving task success rates by 10% compared to traditional RL methods. Through extensive evaluation, we identify key curriculum design parameters—sub-task selection, transition timing, and learning rate—that optimize learning efficiency and provide conceptual guidance for curriculum-based RL frameworks, offering valuable insights into curriculum design in RL and robotic learning.*

## 1. INTRODUCTION

As science and technology advance, the landscape of the engineering work process is continuously extended and revolutionized, which can be initiated and carried out by robotic systems. In the domain of robotic manipulation, machine learning has significantly advanced its capabilities in handling objects and executing complex tasks. Machine learning (ML) has been utilized to develop robotic systems capable of learning to grasp and manipulate objects of varying shapes and sizes, thereby addressing challenges in dexterous manipulation tasks [24]. Furthermore, it has been proposed that robots be adapted to unseen scenarios, enabling self-learning in robots and significantly reducing the need for reprogramming and human intervention [25]. On the other hand, reinforcement learning (RL), as one of the main branches of machine learning, has shown promise in solving the robotic work process by interacting with environments. RL algorithms enable robots to make a series of decisions that optimize their behavior through trial and error, learning from past actions to improve future performance. This method is particularly effective in dynamic and unpredictable environments where pre-programmed behaviors may not suffice [4]. For example, RL has been applied to develop autonomous robotic systems that adapt to changing conditions in real-time, such as adjusting strategies in response to obstacles or system failures [26].

The current technological landscape showcases impressive strides in integrating RL algorithms into operations. Despite these advancements, effectively and reliably applying ML remains a significant challenge. One of the main obstacles is the limited capacity of agents to fully comprehend the intricacies of real-world dynamics and objectives. This gap in understanding hinders agents from achieving the desired level of reliability and adaptability. Addressing these challenges calls for concerted efforts to enable agents' deeper understanding and interpretation of the real world. RL is a potential solution since it allows agents to learn a dynamic environment through interactions. Still, extensive exploration and trial-and-error are often required to develop effective strategies. It can be exceedingly sample-inefficient and demand substantial computational resources when the scenarios are highly complex, making RL not feasible in practice. We often lack a clear understanding of how to assist the agent in achieving the desired robustness in diverse environments because the way the agent learns and the knowledge it learns are uninterpretable, further complicating the application of RL in engineering work processes. There is an urgent need for a comprehensive understanding of the agent learning process and for developing more efficient methods to strategically leverage the learned knowledge and adapt with less training effort.

Enabling the agent to gain a deeper understanding and adapt more efficiently to diverse real-world working scenarios is crucial, and strategic knowledge utilization is a key factor in this process. Knowledge-based engineering (KBE) aims to achieve design automation via knowledge capture. The idea can be applied to the operation process by allowing the agent to capture knowledge during the learning process and then further re-use it in related areas without extensively exploring from scratch [27], significantly enhancing the efficiency and robustness of problem-solving.

To demonstrate the effectiveness of integrating this idea into the RL process, we conduct a case study using the two-block stacking task from the CausalWorld benchmark [1]. Stacking two blocks is a particularly challenging task. While a reward function is provided and a SAC-trained model serves as a baseline, previous research has primarily focused on algorithm development using this predefined reward function. However, the fractional success rate for all approaches, including the baseline model, remains below 50%, even after prolonged training [1, 28]. This suggests the underlying knowledge required for successful task completion is not fully captured. To address this limitation, we propose a refined reward function that removes rigid learning conditions and avoids translating agent learning into human learning paradigms. Instead of enforcing a strict order of learning or requiring perfect conditions, our approach allows the agent to simultaneously optimize multiple movement components. As a result, we observe a significant improvement in the final fractional success rate and learning efficiency compared to the original reward function.

Our previous research exploring CausalWorld [3] found that curriculum learning benefits complex task learning by decomposing tasks and allowing agents to master them step by step. This effect is particularly strong in cross-task curricula, where pre-training on a simple grasping task significantly reduces the training time for subsequent pushing tasks while requiring less pre-training time than an in-task curriculum. Additionally, we observed that sub-task selection and transition timing play critical roles in curriculum learning. Premature transitions hinder the agent's knowledge acquisition, while excessive delays lead to overfitting on more straightforward tasks. Building on these insights, we extend our study from a single-block task to a two-block assembly task, where the agent must distinguish between blocks and operate with higher precision to avoid collisions and successfully construct the assembly. Furthermore, we extend manipulation complexity from 2D to 3D movement, incorporating heterogeneous movements to further challenge the agent.

To address these increasing complexities, as well as fully understand the agent's learning process and enhance the learning efficiency through strategically re-use the knowledge,, we propose the KCAC framework, which integrates knowledge transfer into reinforcement learning. This framework provides a comprehensive engineering design, defines an across-task curriculum, and identifies key design parameters, including sub-task selection, transition timing, and learning rate. We further analyze their impact on learning efficiency and task performance.

## 2. RELATED WORK
### 2.1 Reinforcement Learning

Engineering design is a multidisciplinary topic that requires careful optimization to balance performance and efficiency across various constraints [8, 9]. Reinforcement learning is one of the solutions, allowing agents to understand and optimize the process through continuous interaction with their environment [14-16].

Reward function design is pivotal in enhancing the effectiveness of robotic learning applications. RL agents learn optimal behaviors guided by reward signals that indicate the desirability of specific actions. Multi-objective reward function enables agents to learn policies that consider various objectives simultaneously. Kaushik et al. introduced a multi-objective model-based policy search algorithm that addresses sparse rewards by optimizing multiple objectives. Their approach maximizes the expected return and significantly reduces interaction time in a simulated robotic arm application [17]. For complex compositional robotic tasks, a compound reward function is necessary. Dense rewards provide informative signals to accelerate learning convergence, while sparse rewards define specific sub-goals. Rati Devidze et al. proposed a novel framework for reward function design that balances informativeness and sparseness, ensuring that the learned policy aligns with the intended target behavior [19]. Inspired by these studies, we decompose the problem in our case study into sub-goals and adopt a compound reward function with multi-objective optimization. Sparse rewards guide the agent toward achieving sub-goals, while dense rewards provide continuous feedback, encouraging the agent to follow the desired trajectory.

For tasks of high complexity, which are challenging to address using traditional reinforcement learning alone, a curriculum learning strategy has been employed to enhance the learning process [20]. It involves decomposing a task into solvable sub-tasks and allows for a gradual and systematic approach to tackling complex tasks. The learning process is heavily affected by the curriculum design, which becomes the main challenge. Narvekar et al. proposed a framework for effective curriculum generation, focusing on the creation of source tasks by leveraging domain knowledge and observations of the agent's performance [21]. Ryu et al. introduced an approach that utilizes large language models to automatically design task curricula, significantly reducing reliance on extensive domain expertise and manual intervention [22]. Tzannetos et al. developed a curriculum strategy that selects tasks based on difficulty, ensuring they are neither too easy nor too hard. Their method aims to accelerate RL training while minimizing dependence on domain-specific knowledge and hyperparameter tuning [23].

Regardless RL enables agents to learn in dynamic environments through interactions, it often requires extensive exploration of state-action combinations, leading to low sample efficiency, which can be impractical in real-world applications. Additionally, the nature of RL models makes it challenging to understand the reasons behind training failures and to propose corresponding solutions [12, 13]. This lack of interpretability hinders the deployment of RL in critical applications. Therefore, research focused on enhancing sample efficiency and developing explainable RL algorithms is essential to facilitate the practical adoption of RL in real-world scenarios.

## 2.2 Knowledge-Based Engineering

Enabling the agent to gain a deeper understanding and adapt more efficiently to diverse real-world working scenarios is crucial, and strategic knowledge utilization is a key factor in this process. Knowledge-Based Engineering (KBE), aiming to achieve design automation via knowledge capture, can be potentially applied to the operation process. It allows the agent to capture knowledge during the learning process for further re-use in related areas without exploring extensively from the very beginning, significantly enhancing the efficiency and robustness of problem-solving.

Knowledge within the context of learning and adaptation can be classified broadly into two categories: explicit and implicit. Explicit knowledge is structured information that can be easily documented, shared, and codified. It's the kind of knowledge that is found in tangible forms like books, databases, manuals, or procedures [29]. In contrast, implicit knowledge is often not consciously recognized. It is the tacit insight that is shared through illustration or the performance of tasks and can be transferred from one task to another. Features or patterns are examples of implicit knowledge, which can be discerned by learning models [30]. KBE is effective for explicit knowledge capture. For example, the ICARE model, standing for Illustrations, Constraints, Activities, Rules, and Entities, was leveraged in widely used KBE methodologies to decompose knowledge elements and create a problem representation from multiple views. Then, the knowledge is formalized to be acceptable to engineers [27]. The Design and Engineering Engine (DEE) was proposed to reduce the expenses of the repetitive and non-added-value activities for multidisciplinary design processes and support creative and interactive design. It incorporated with KBE, the toolboxes within the DEE framework represented various conceptual and preliminary design methods, enabling comprehensive generation and analysis capability [31].

However, the explicit knowledge capture proved to be case-based and ad hoc, leading to inefficiencies in engineering product progress or when condition or design goals shift. Knowledge loss was also a major issue due to the difficulty of knowledge re-use [32]. As mentioned by Verhagen et al., instead of defining rule-based knowledge, automatic knowledge extracted from applications is needed. By learning that kind of knowledge, designers, and engineers can update models and re-use the knowledge. Machine learning models are utilized to encode patterns for prediction and generation, which can be especially beneficial in applications that require extracting insights from data for informed decision-making [5, 11, 33, 34]. For example, a model initially trained for broad image classification on a large dataset can serve as the foundational architecture for more specialized applications, such as manufacturing failure analysis based on sensory data prediction [35] and image recognition in self-driving vehicles [7]. Despite these advancements in single-procedure tasks, knowledge capture in work processes that involve a systematic sequence of steps, such as autopiloting to a designated goal or assembling complex structures—remains underdeveloped. This gap highlights the ongoing challenge of devising appropriate methodologies and frameworks to effectively capture and utilize knowledge in multi-step processes.

## 3. METHOD

### 3.1 Re-design the Baseline Reward Function

The task used in this paper to evaluate our proposed framework is illustrated in **Figure 1**. The robot arm, equipped with three fingers controlled by three actuators, manipulates a central block (shown in red) toward its goal positions (depicted in green). The objective is to stack the central block on top of another one while keeping the bottom block stationary. The task is considered successful when the bottom of the red blocks (objects) overlap entirely with the top of the green blocks (goals), achieving the desired final state. Reinforcement learning is used for agent learning. The state input is a 56-dimensional array, which includes task-relevant variables such as the remaining time for the task, joint positions, joint velocities, and the linear velocities of blocks, among others. The action output is a 9-dimensional array corresponding to the joint positions of each finger, sampled from a predefined continuous space.

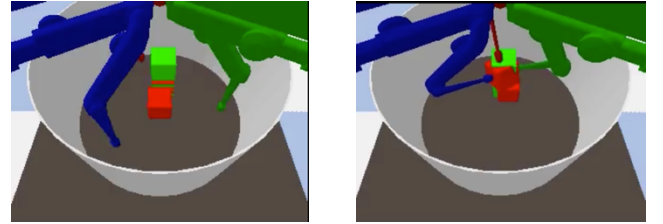

(a) Initial state  (b) Final state

**FIGURE 1:** Illustration of stacking task

We define these elements as follows:

- block_1: The stationary bottom block
- block_2: The block to be stacked on top
- goal_block_1: The desired position of the bottom block
- goal_block_2: The desired position of the top block

As previously mentioned, we hypothesize that the primary factor limiting the performance of baseline models and related research is the *reward function* rather than deficiencies in reinforcement learning or transfer learning algorithms. The benchmark reward function proposed in [1] is defined as follows.

$$R_{stack} = \mathbf{1}_{d^t(o_1,e)>0.02}(-750\Delta^t(o_1,e) - 250\Delta^t(o_1,g_1)) + \mathbf{1}_{d^t(o_1,e)<0.02}\left(-750\Delta^t(o_2,e) - 250(|o_{2,z}^t - g_{2,z}^t| - |o_{2,z}^{t-1} - g_{2,z}^t|) - \mathbf{1}_{o_{2,z}^t - g_{2,z}^t>0}125\Delta^t(o_{2,x,y}, g_{2,x,y})\right) + R_{goal_1} + R_{goal_2} + 0.005 * \|v^t - v^{t-1}\| \quad (1)$$

where $e$, $o$, $g$, and $v$ represent end-effector position, block position, goal block position, and joint velocity, respectively, and $d^t$ represents the distance between 2 objects at time $t$.

The reward function used in the experiment is a compound function, defined as a conditional sum of multiple components. $\Delta^t$ is a dense reward, representing the distance difference from the previous timestep. The positive reward is given if the end-effector gets closer to the blocks or the blocks get closer to the goal position, which can be useful for reward shaping in reinforcement learning. $R_{goal_1}$ and $R_{goal_2}$ are sparse rewards, defined in equation (2). $A_{intersect}$ represented the intersection area between the current block position and the desired goal position while $A_{union}$ representing the union of the two positions. It becomes greater than 0 when the block intersects with the goal, guiding the agent to learn the assembly task.

$$R_{goal} = \frac{A_{intersect}}{A_{union}} \quad (2)$$

The function is designed to guide a stacking task through different stages. Each component is activated by an indicator function $\mathbf{1}_{condition}$, which equals 1 when the condition is satisfied and 0 otherwise. For instance, the first condition indicates when the distance between the end-effector and the first block $d^t(o_1, e)$ is greater than 0.02, the reward encourages minimizing both the distance between the end-effector and the block $\Delta^t(o_1, e)$ and the distance between the block and its goal position $\Delta^t(o_1, g_1)$). Once the end-effector is sufficiently close to the first block ($d^t(o_1, e) < 0.02$), the reward shifts to the second condition to focus on the second block. $\Delta^t$ is a dense reward, representing the distance difference from the previous timestep. The positive reward is given if the end-effector gets closer to the blocks or the blocks get closer to the goal position, which can be useful for reward shaping in reinforcement learning.

To simplify the task and reduce training time, we initialize block_1 at the goal position (goal_block_1). This deactivates the first condition, while the second condition is active from the start, meaning the agent only needs to learn grasping, lifting, and placing block_2 onto block_1. Upon implementing the baseline model and evaluating its performance through metrics and training videos, we observed that the agent reaches a local optimum, where it correctly places block_1 but neglects block_2.

The first condition activates only the reward for block_1 if its distance from the goal exceeds 0.02, while the reward for block_2 remains zero. This forces the agent to ignore block_2 until block_1 is correctly placed. The *second condition* enables the reward for block_2 only when block_1 is correctly positioned, restricting early-stage learning. Even though block_1 is initialized at the goal, it can easily be displaced before the agent has mastered the task. This raises the question: *Why can't the agent learn to pick up block_2 first, even if it slightly touches block_1*? The grasping behavior can be fine-tuned later to prevent disturbing block_1. Instead of imposing rigid constraints, the reward function should encourage the agent early rather than requiring perfect actions to earn positive rewards. The *third condition* enforces an undesirable learning order, requiring the agent to lift block_2 higher than goal_block_2 before activating the horizontal movement reward. However, this contradicts an efficient motion strategy: *Why can't the agent learn a curved movement, performing vertical and horizontal motion simultaneously?*

Based on our hypothesis, the existing reward structure imposes an unnatural learning sequence, which should be eliminated. Instead of restricting the agent's learning order, the reward function should maximize the total reward and flexibly encourage task completion. Thus, we removed the "if conditions" and applied reward engineering to develop a more effective reward function, shown below.

$$R_{stack} = -750\Delta^t(o_2, e) - 250(|o_{2,z}^t - g_{2,z}^t| - |o_{2,z}^{t-1} - g_{2,z}^t|) - 125\Delta^t(o_{2,x,y}, g_{2,x,y}) + 0.5 * R_{goal_1} + 1 * R_{goal_2} + 0.005 * \|v^t - v^{t-1}\| \quad (3)$$

We adjust the weights of $R_{goal_1}$ and $R_{goal_2}$ to emphasize block_2 manipulation, guiding the agent to focus on lifting and placing block_2. We remove the reward for grasping and moving block_1 as it is already initialized at goal_position_1 and $R_{goal_1}$ sufficiently ensures block_1 aligns with its desired position. Our revised approach successfully improves the agent's ability to complete the stacking task efficiently.

### 3.2 KCAC Framework

To enhance efficiency and interpretability in reinforcement learning (RL) for complex work processes, we propose a novel modeling architecture—the KCAC system (Knowledge Capture, Adaptation, and Composition). KCAC integrates knowledge engineering into reinforcement learning, enabling more structured learning and transferability. Designed for flexibility and broad applicability, KCAC is compatible with various RL algorithms and can be adapted to different applications.

KCAC is built upon fundamental RL principles, incorporating current and subsequent states $s$, $s'$, action $a$, action-value function $Q(s, a; \theta)$ with parameter $\theta$, reward $r$, and policy $\pi(a|s; \theta)$. To effectively ***capture*** knowledge in complex engineering environments, KCAC employs a compound reward function design, as introduced in Section 3.1. Furthermore, the framework establishes a systematic transfer learning process, allowing the agent to ***adapt*** knowledge from sub-tasks and ***compose*** complex target tasks using a novel parameterization strategy that enhances learning efficiency.

As the KCAC procedure outlined in **Table 1**, function $G(S)$ supports curriculum generation. It decomposes the current target task $S$ into a series of sub-tasks $S_1$, $S_2$, ..., $S_N$, with increasing complexity based on the reward function components of the target task. Together, these sub-tasks form a curriculum for learning the target task, allowing the agent to progressively acquire knowledge before tackling the final objective. The last sub-task $S_N$ is identical to the original target task $S$. Based on the similarity between sub-tasks and the target task, the function $M(<S_N>)$ is defined to guide the selection of a set of optimal knowledge transition timings $<T_N>$, and learning parameters $m$ including the learning rate, entropy, and target network update coefficient for new sub-task learning.

**Table 1.** Procedure of KCAC

---
Generate sub-tasks:

$G(S) \rightarrow <S_N>$

Generate transition timing and learning rate:

$M(<S_N>) \rightarrow <T_N>, m$

Initialize $i = 1$

For number of episodes $t$ from 1 to $T_N$, do

    If $t == T_i$:

        Perform knowledge transfer: $\theta = \theta_{s_i}$

        $i \mathrel{+}= 1$ if $i < N$

    Initialize state $s$

    While not terminal state:

        Select action a using policy $\pi(a|s; \theta)$

        Execute $a$, observe $r$ and $s'$

        Update $\theta$ by $\nabla L(\theta)$ with $m$

        Update current state $s \leftarrow s'$

---

In the learning process, after sub-tasks curriculum and learning parameter generation, the agent begins learning from the first sub-task, progressing episode by episode while interacting with the environment as in a standard reinforcement learning process. Upon reaching the transition timing for the next sub-task, the agent transfers previously learned weight and adjusts learning parameters based on $m$. This process continues until the agent learns the final target task $S_N$. The detail of $G(S)$, $M(<S_N>)$ will be explained in Section 4, using the stacking of two blocks as a case study.

## 4. ROBOTIC MANIPULATION TASK ENGINEERING DESIGN

In this section, the KCAC framework is constructed and evaluated using the task of stacking two blocks. The performance of the redesigned reward function is compared with the original reward function to verify the effectiveness of the improved reward design. Next, knowledge transfer is introduced and compared with direct learning using the improved reward function to demonstrate higher training efficiency and provide insights into design parameter selection. To facilitate learning the two-block stacking task, we designed a two-stage cross-task curriculum using grasping and picking tasks. The agent first undergoes pre-training on grasping or picking before transitioning to the stacking task at an optimized transition point. During this transition, the agent transfers previously learned weights and initializes learning parameters for stacking.

Our previous research [3] highlights the critical role of transition timing in performance. Additionally, the similarity between the source and target tasks significantly influences knowledge transfer efficiency [6, 10, 11]. To further investigate these factors, we examine the effects of transition timing and learning parameters across multiple two-stage learning curricula with varying sub-task similarity levels. Moreover, experimental observations indicate that the default learning rate provided in the baseline is not optimal for all curricula with different task similarities. As a result, the learning rate was also varied and considered a key design parameter in this study.

Based on these findings, the function $M(<S_N>)$ is conceptually defined to determine optimal transition timing and learning parameter selection. 3-stage curriculums (grasping, picking and stacking) are developed using prior conclusions to further improve the learning efficiency. The results are compared with 2-stage curriculum candidates, leading to the conceptual definition of function $G(S)$, which contributes to curriculum generation.

### 4.1 Sub-tasks Generation

From **Equation (3),** the success of the stacking task requires the end-effector to approach block_2, grasp it securely, lift it, and place it on top of block_1. Mastering the early-stage skills of grasping and lifting block_2 can significantly benefit learning the full stacking task. Therefore, we introduce grasping and picking as sub-tasks to facilitate curriculum generation.

As shown in **Figures 2** and **3**, the goal of the grasping task is to position the robot's three fingers around block_2 without displacing it or colliding with block_1, as illustrated in **Figure 2**. The objective of the picking task is to grasp block_2 and lift it vertically without any horizontal movement while avoiding collisions with block_1. As shown in **Figure 3**, block_2 should fully overlap with goal_block_2 at the correct height in the stacking task.

The reward functions for these two sub-tasks are derived from the stacking task reward function and are formally defined in **Equations (4)** and **(5).** Based on experimental results and reward engineering, the grasping task employs a slightly different reward function design than stacking task. The dense reward for distance from previous time step is based on the distance from initial position of block_2, preventing block_2 to be moved away during grasping. The sparse reward is based on the distance between the end-effectors and the block. This design accelerates learning by encouraging efficient grasping strategies.

$$R_{grasp} = -750\Delta^t(o_2, e) - 250\Delta^t(|o_2^t - o_{2\_init}^t| - |o_2^{t-1} - o_{2\_init}^t|) + 0.5 * R_{goal_1} + min(1, 1 - \frac{d^t(o_2,e) - min\_dist}{max\_dist - min\_dist}) \quad (4)$$

$$R_{pick} = -750\Delta^t(o_2, e) - 250(|o_{2,z}^t - g_{2,z}^t| - |o_{2,z}^{t-1} - g_{2,z}^t|) + 0.5 * R_{goal_1} + 1 * R_{goal_2} \quad (5)$$

5                                                                  © 2025 by ASME

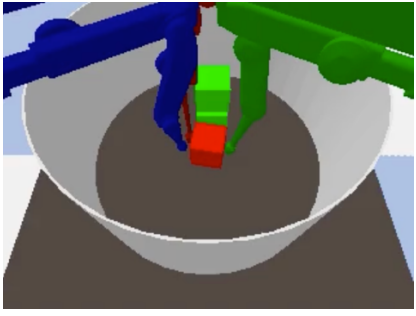

**FIGURE 2:** Illustration of grasping task

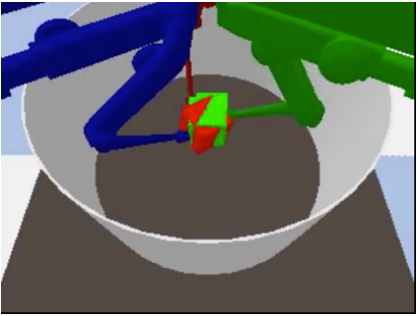

**FIGURE 3:** Illustration of picking task

## 4.2 Similarity Measurement

The relationship between reward functions is used to compute the similarity between sub-tasks and the target task. Intuitively, one might assess similarity based on the weights of shared and unshared reward components in the compound reward function. However, the scales of different reward components vary significantly. For instance, the distance between end-effectors and blocks is typically around 1e-3, while the intersection between the current and desired goal positions is around 1e-1. Consequently, directly using the raw coefficients of components for similarity computation may be misleading. After monitoring each component, we observed that when the task is appropriately implemented, the relative contributions of these components remain comparable. Therefore, we represent each task as a binary vector, where each reward term is either present (1) or absent (0). This approach allows us to focus on the structural presence of reward components rather than their magnitudes, leading to a more robust evaluation of task similarity.

To systematically analyze task similarities, we first identify the key reward components present in each task, including temporal distance difference and sparse goal reward. Then, the reward function of all tasks is expressed in a general form as follows:

$$R = R_{end} + R_{move} + R_{vert} + R_{goal_1} + R_{goal_2} + R_{dist} + R_{hori} + R_{vel} \qquad (6)$$

where each term represents a specific aspect of the task.

- $R_{end}$: Movement of end-effectors to block_2
- $R_{move}$: Movement of block_2 from its initial positions
- $R_{vert}$: Vertical movement of block_2 to goal height
- $R_{goal_1}$: Alignment with the goal_block_1
- $R_{goal_2}$: Alignment with the goal_block_2
- $R_{dist}$: Distance from end-effectors to block_2
- $R_{hori}$: Horizontal movement of block_2
- $R_{vel}$: End-effectors moving velocity

Each task is then represented as a binary vector, where the presence of a reward term is denoted as 1 and absence as 0. The resulting vectors are, grasping task $V_G = [1,1,0,1,0,1,0,0]$, picking task $V_P = [1,0,1,1,1,0,0,0]$, stacking task $V_S = [1,0,1,1,1,0,1,1]$.

Using cosine similarity, defined as:

$$Sim(V_A, V_B) = \frac{V_A \cdot V_B}{\|V_A\| \|V_B\|} \qquad (7)$$

we compute the similarities:

- Grasping and stacking task: $Sim(V_G, V_S) = 0.4$
- Grasping and picking task: $Sim(V_G, V_P) = 0.5$
- Picking and stacking task: $Sim(V_P, V_S) = 0.8$

These results confirm our expectations: grasping and stacking share low similarity, grasping and picking exhibit medium similarity, while picking and stacking are highly similar. This binary representation allows us to evaluate task relationships in a structured manner, focusing on the presence of reward components rather than their varying magnitudes.

## 4.3 Reinforcement Learning Settings

The experiment is implemented using Soft Actor Critics from Stable baseline 3. Each result is generated and averaged from 5 random seed. 3 set of learning parameter are used to discover the impact, as shown in Table 2. In this session, we use lr_1e-4, lr_5e-5, lr_1e-5 to represent these 3 sets of parameters respectively.

The experiment is conducted using the Soft Actor-Critic (SAC) algorithm [2], implemented in PyTorch via the Stable-Baselines3 library. Each result is obtained by averaging over five runs with different random seeds. To analyze the impact of learning parameters, we evaluate three different learning rate settings with corresponding adjustments of other hyper-parameters, as detailed in Table 2. For clarity, we denote these settings as lr_1e-4, lr_5e-5, and lr_1e-5, respectively. The default learning rate of 1e-4, as provided by the baseline, is used for all pre-training phases in the curriculum.

Table 2. Learning parameter

| denotation | learning rate | tau | entropy coeff | batch size | buffer size | discount | target entropy |
|---|---|---|---|---|---|---|---|
| lr_1e-4 | 1e-4 | 1e-3 | 1e-3 | 256*4 | 1e6 | 0.95 | auto |
| lr_5e-5 | 5e-5 | 1e-4 | 1e-4 | 256 | 1e6 | 0.95 | auto |
| lr_1e-5 | 1e-5 | 1e-4 | 1e-4 | 256 | 1e6 | 0.95 | auto |

## 5. RESULTS AND DISCUSSION

### 5.1 Impact of Reward Function Design

We redesign the reward function so the scale of the reward of baseline and our model become different. We cannot compare the performance by episodic reward during the learning process like other RL studies did. Instead, we compare the fractional success, $\frac{A_{intersect}}{A_{union}}$, as introduced in **Session 3.1**. The significant improvement of overall fractional success and top block fractional success of our refined model compared to the baseline model are shown in **Figures 4** and **5**.

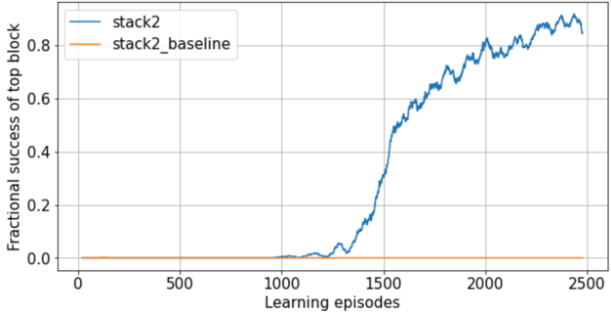

**FIGURE 4:** Fractional successful rate of top block

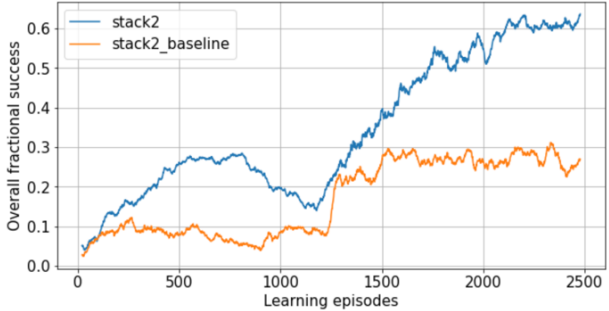

**FIGURE 5:** Fractional successful rate of two blocks

From **Figure 4**, we can see that the top block has not been activated in the baseline, which is the main reason for the poor performance of the baseline model. Our model significantly improves the top block's fractional success regardless of the conditions and causing learning sequence. In the following session, we will continue to use the fractional success of the top block as the evaluation metric to evaluate the further improvement due to the cross-task curriculum learning.

We hypothesize that the stacking task requires heterogeneous movements, including grasping the block, moving vertically and horizontally, and stabilizing the assembly. Each movement consists of a sequence of actions, primarily defined by the end-effector's positional changes, as described in the original paper [1]. Unlike humans, reinforcement learning agents do not inherently discover an optimal learning order for heterogeneous movements through environment interactions, even when guided by conditional rewards. RL agents aim to generate a policy that selects actions to maximize total reward rather than adhering to a predefined learning sequence. However, the conditional statements in the original reward function impose a *sequential learning order*, which hinders learning efficiency.

Instead of restricting the agent's learning progression, the reward function should be designed so that reward components can be incrementally optimized even in early training stages, allowing the agent to *maximize total reward simultaneously* and encourage task completion more flexibly. This is the primary reason behind the effectiveness of our refined reward function design. By removing restrictive conditional dependencies, our design ensures that reward signals remain informative even in early stages, reinforcing incremental progress rather than delaying positive feedback until specific conditions are met. This enables the agent to optimize all movement components gradually, leading to faster convergence and better overall task performance.

### 5.2 2-Stage Learning: Impact of Knowledge Transfer Parameters

To facilitate learning the two-block stacking task, we designed a 2-stage cross-task curriculum using grasping and picking tasks. We explore two curriculum variations: grasping-to-stacking and picking-to-stacking. Each has distinct advantages. The *grasping-to-stacking* curriculum provides an easier starting point, making pretraining more straightforward. However, due to the low similarity between grasping and stacking, additional learning time is required in the second stage. Conversely, the *picking-to-stacking* curriculum presents a more challenging starting point, as the picking task requires longer pretraining. However, since picking and stacking are highly similar, the transition to stacking is smoother, requiring less adaptation. Additionally, we investigate a grasping-to-picking curriculum with medium similarity, which serves as a foundation for a 3-stage curriculum. This allows us to analyze the combined effects of similarity, transition timing, and learning rate.

The results are shown in **Figures 6–14**. In these plots, the blue line labeled "stack2" represents the fractional success rate of directly learning the two-block stacking task using our refined reward function. Other lines represent different two-stage curriculum learning approaches, where knowledge transfer is integrated into reinforcement learning. The same reward function is used across all experiments, and we analyze the impact of various *knowledge transfer parameters*. The legend annotations indicate specific curriculum settings. For example, grasp2_60_curri_lr_1e-5 represents a grasping-to-stacking curriculum where the agent is first pretrained on the grasping task using the default learning rate for 60 episodes before transitioning to the stacking task with a learning rate of 1e-5, as detailed in **Table 2**. Note that only the last stage learning (stacking task) is shown in the figures since the fractional success of target tasks cannot evaluate previous stage tasks.

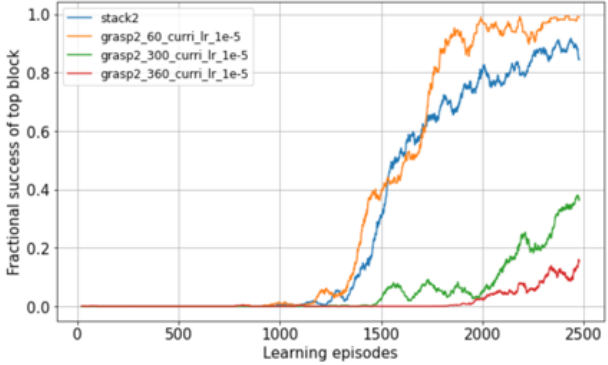

**FIGURE 6:** Grasping-stacking curriculum with lr_1e-5 learning process

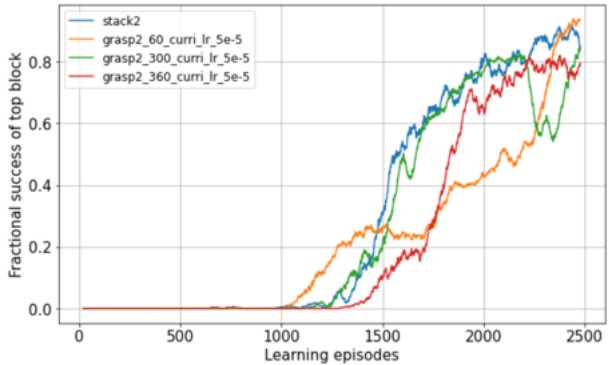

**FIGURE 7:** Grasping-stacking curriculum with lr_5e-5 learning process

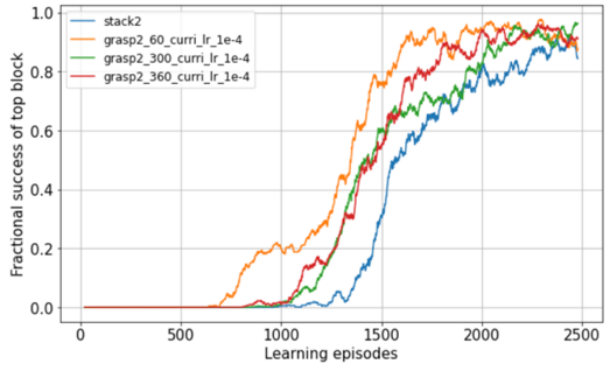

**FIGURE 8:** Grasping-stacking curriculum with lr_1e-4 learning process

From all the figures, we observe that learning performance is highly sensitive to the ***learning rate***. While both the high learning rate (1e-4) and the low learning rate (1e-5) yield workable results depending on task similarity, the medium learning rate (5e-5) proves to be unstable across all cases. This highlights the critical importance of selecting an appropriate learning rate.

Grasping and stacking share the ***lowest similarity*** among the examined tasks. As shown in **Figures 6–8,** the highest learning rate (1e-4) results in the best performance, with all curriculum variations outperforming direct RL learning. Regardless of the learning rate, a 60-episode pretraining phase on the grasping task enables the fastest convergence on stacking and achieves a higher final fractional success compared to longer pretraining durations or direct RL. Despite the low similarity, pretraining still benefits the target task. In this case, an early transition and a higher learning rate are crucial—full mastery of the pretraining task is unnecessary, and faster adaptation is key for low-similarity curricula.

For the grasping-to-picking curriculum, which exhibits ***medium similarity*** (**Figures 9–11**), a higher learning rate (1e-4) is again preferable, facilitating a faster increase in fractional success than direct RL. However, as task similarity increases, the earliest transition timing is no longer optimal. The transition should occur later to allow the agent to develop a better grasp of the pretraining task, as these skills are more transferable to the target task. However, delaying too much can be detrimental—**Figures 9–11** show that a 360-episode pretraining phase results in lower performance than 300 episodes. This aligns with our previous research [3], which found that *late transitions lead to overfitting* on the *pretraining task*, thereby hindering adaptation to the target task. Our findings suggest 300 episodes strike the right balance, ensuring sufficient training in the first stage while mitigating overfitting. This observation will be further explored in the subsequent three-stage curriculum design.

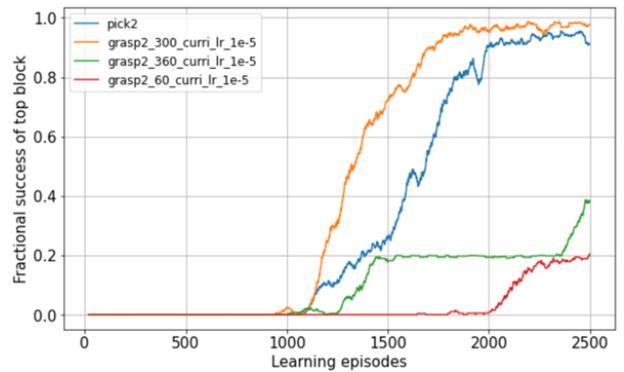

**FIGURE 9:** Grasping-picking curriculum with lr_1e-5 learning process

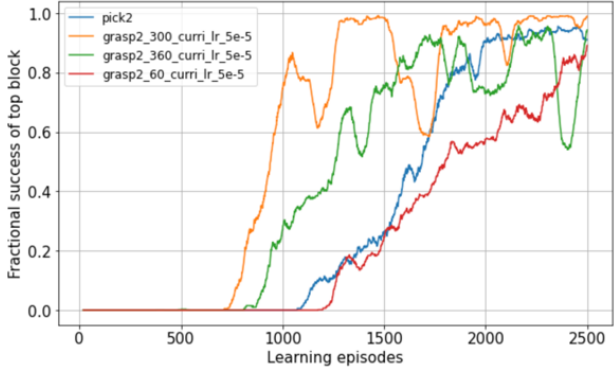

**FIGURE 10:** Grasping-picking curriculum with lr_5e-5 learning process

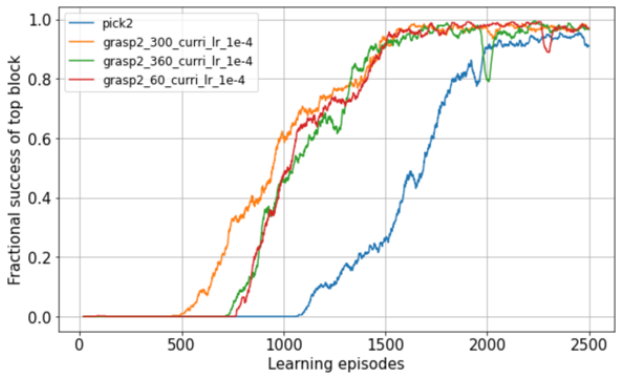

**FIGURE 11:** Grasping-picking curriculum with lr_1e-4 learning process

For the picking-to-grasping curriculum, a longer pretraining phase is required because the picking task involves vertical movement and goal shape alignment, making it more challenging to learn compared to the grasping task used in previous curricula. In *high-similarity* curricula, the learning content of the two tasks significantly overlaps, particularly in the early stages (grasp and lift). The overfitting issue of late transition in the medium-similarity case is less pronounced here, as the high similarity between tasks ensures smoother knowledge transfer. Consequently, the knowledge transfer process is less sensitive to transition timing than the medium-similarity case.

As shown in **Figures 12–14**, a pretraining duration of 1800 episodes yields slightly better results than 1200 episodes. *Longer pre-training time* leads to quicker adaptation. However, from the figure, we can see that the extra 600 episodes of training only leads to less than 100 sooner convergence; the trade-off is not computing resource friendly since the pre-training effort also needs to be taken into consideration for curriculum design. For learning rate analysis, a lower learning rate (1e-5) facilitates an earlier increase in the success rate and quicker convergence, which is expected in high-similarity cases. Since most of the relevant knowledge is already covered in the first stage, only fine-tuning with a low learning rate is required in the second stage. Comparing **Figures 12** and **14**, we observe that a higher learning rate modifies the model's weights too drastically, delaying the initial increase in success rate from zero.

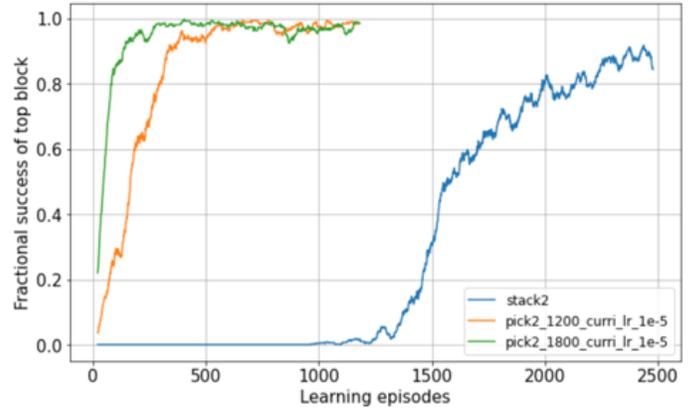

**FIGURE 12:** Picking-stacking curriculum with lr_1e-5 learning process

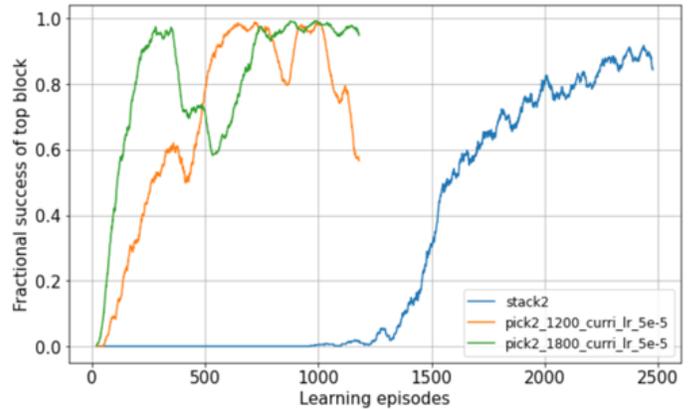

**FIGURE 13:** Picking-stacking curriculum with lr_5e-5 learning process

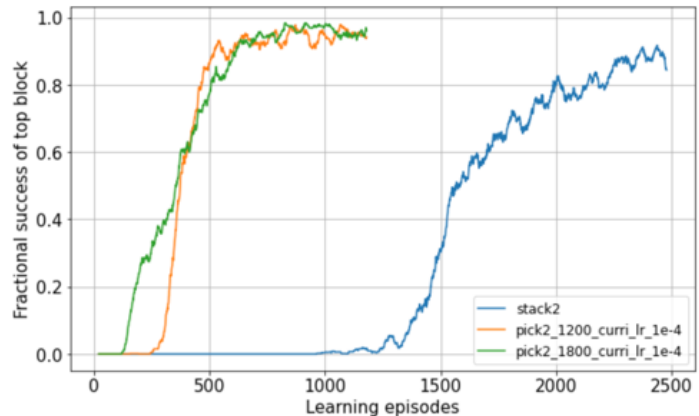

**FIGURE 14:** Picking-stacking curriculum with lr_1e-4 learning process

## 5.3 3-Stage Learning: Design an Efficient Curriculum

To accelerate the learning process of the stacking task, we design a 3-stage curriculum: ***grasping to picking to stacking***, building on insights from 2-stage curriculum learning. Compared to the most efficient 2-stage curriculum (picking to stacking), adding the grasping task is expected to shorten the required picking pretraining time, ultimately benefiting the final stacking task.

For the first stage, as demonstrated in **Section 5.2**, 300 episodes of grasping serve as an optimal transition point for picking task learning, making it a selected variation for the three-stage curriculum. Additionally, 60 episodes of grasping have also shown benefits in picking with much less pre-training time, so this setting is included for comparison. For the second stage, based on **Figure 11**, 900 episodes of picking training with prior grasping pretraining achieve performance equivalent to 1800 episodes of direct picking learning. 1800 episodes of picking pretraining have been shown to provide the most significant boost to stacking task learning. Thus, 900 episodes of picking are applied for the second stage of the three-stage curriculum.

In **Figure 15**, we compare the learning process of the 3-stage curriculum against the best-performing 2-stage setup, picking to stacking with a learning rate of 1e-5, as shown in **Figure 12**. While the **2-stage** learning approach with 1800 episodes of picking pretraining achieves the fastest learning (orange line), it requires significantly more pretraining time, making it less efficient than the 3-stage approach, which achieves comparable performance with equal or less than 1200 episodes of pretraining. For the **3-stage** curriculum (red line), the total pretraining time consists of 300 episodes of grasping followed by 900 episodes of picking, summing to 1200 episodes. This setup results in a noticeable jump-start and faster convergence compared to the two-stage learning approach with the same pretraining time (green line), demonstrating the ***effectiveness of the 3-stage curriculum***. Additionally, the variant with only 60 episodes of grasping pretraining (purple line) eventually catches up, requiring even less pretraining while achieving a higher final fractional success rate.

Overall, for all successful curriculum learning candidates, the convergence time for the stacking task is approximately 540 episodes. The total learning time includes both pretraining and target task learning. The best-performing candidate is the 3-stage curriculum with 60 episodes of grasping pretraining (purple line), which completes training in 60 + 900 + 540 = 1500 episodes. This represents a *significant 40% reduction in training time and 10% success rate improvement* compared to direct stacking learning (blue line), which requires 2500 episodes and still fails to achieve 100% fractional success. These results demonstrate the efficiency and effectiveness of the KCAC framework in optimizing reinforcement learning through structured curriculum design.

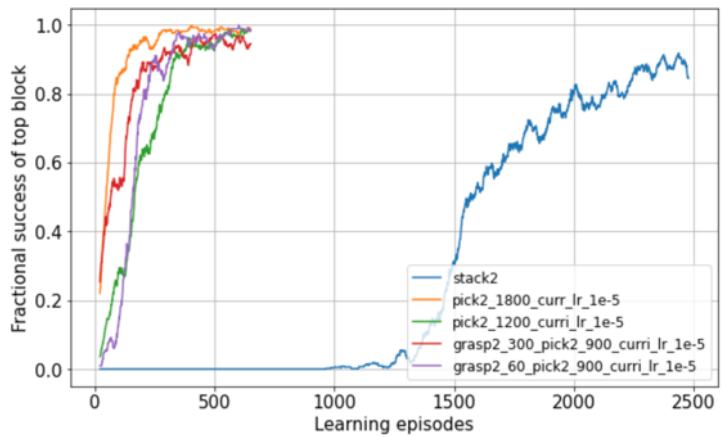

**FIGURE 15:** Comparision of 2-stage learning and 3-stage learning

## 6. CONCLUSION AND FUTURE WORK

In this research, we use the two-block stacking task as a case study, extending our investigation from a single-block task to a more complex two-block assembly task. This task requires the agent to differentiate between blocks and operate with higher precision to avoid collisions while successfully constructing the assembly. Furthermore, the manipulation complexity increases from 2D to 3D movement, incorporating heterogeneous movements to further challenge the robotic agent. To the best of our knowledge, the baseline model and existing RL approaches do not show a high success rate [1, 28], indicating poor knowledge capture in this domain.

To address this challenge and enable transferability to other complex learning-based robotic applications, we propose a reward function design that avoids rigid constraints and does not rely on mapping agent behavior directly to human learning paradigms. This flexible design significantly improves both learning efficiency and final task success rates. To further support effective learning, we introduce the KCAC framework, which integrates knowledge engineering into reinforcement learning through cross-task curriculum learning. This structured approach enables the agent to progressively acquire the stacking task. We also develop a similarity measurement method to quantify the relationship between sub-tasks, guiding curriculum construction. Following is an outline of the key insights gained from the reported research. These contributions aim not only to improve robotic performance on the two-block task but also to inform general principles for knowledge-guided learning in other real-world applications requiring precise control and adaptive decision-making.

- For knowledge **capture**, the reward function should avoid enforcing a strict learning order or requiring perfect execution conditions. Instead, it should allow agents to optimize multiple movement components simultaneously to accelerate the learning process.

- For knowledge **adaption**, by analyzing 2-stage curricula with parameter variations, we derive insights into the design of the function $M(<S_N>)$:
    - Low similarity: Requires a high learning rate and early transition timing to enable faster adaptation.
    - Medium similarity: Exhibits an optimal transition timing that balances knowledge retention and adaptation. A high learning rate is preferable.
    - High similarity: A low learning rate is required to perform fine-tuning rather than significantly altering the model. Longer pretraining improves adaptation speed, but computational efficiency must also be considered when designing a training pipeline.
- For knowledge **composition**, guidance on the conceptual design of $G(S)$ is provided based on the curriculum comparison. In this specific task, 3-stage curricula outperform 2-stage curricula, as each sub-task builds upon the previous one, reducing learning time in later stages.

In our future work, more sub-tasks with varying complexity can be introduced based on a compound reward function. For example, subtasks such as picking a block and holding it statically or picking a block with controlled horizontal movement can be incorporated. Expanding the curriculum into additional stages can provide further guidance for $G(S)$. More finely defined sub-tasks will generate additional similarity data points, assisting in the mathematical formulation of $M(<S_N>)$. Proper transition timing and learning rate settings can be derived based on task-pair similarity, further optimizing knowledge transfer. Lastly, this framework can be generalized and applied to other engineering tasks, broadening its applicability beyond robotics.

## ACKNOWLEDGEMENTS
We adapted the CausalWorld code in various ways described in the paper. The code and implementations can be provided by the corresponding author upon reasonable request.

This paper is based on the work supported in part by the Autonomous Ship Consortium (ASC) with members of BEMAC Corporation, ClassNK, MTI Co. Ltd., Nihon Shipyard Co. (NSY), Tokyo KEIKI Inc., and National Maritime Research Institute of Japan. The authors are grateful for their support and collaboration on this research.

The authors acknowledge the Center for Advanced Research Computing (CARC) at the University of Southern California for providing computing resources for the research reported in this paper. https://carc.usc.edu.